\documentclass[letterpaper, 10 pt, conference]{ieeeconf}  

\IEEEoverridecommandlockouts                              

\usepackage{multirow}
\usepackage{graphicx}
\usepackage{caption}
\usepackage{subcaption}
\usepackage{dashrule}
\usepackage{varwidth}
\usepackage{amsmath}
\usepackage{amssymb}
\usepackage{booktabs}
\usepackage[table]{xcolor}
\usepackage{soul}
\usepackage{siunitx}
\usepackage[most]{tcolorbox}

\title{\LARGE \bf
DiCE-CIR: Direct Composition Learning for Efficient Zero-Shot Composed Image Retrieval
}

\author{Gwang-Ho Na$^{1}$, Ho-Joong Kim$^{1}$, and Seong-Whan Lee$^{1}$%
\thanks{*This work was supported by the Institute of Information \& Communications Technology Planning \& Evaluation (IITP) grant funded by the Korea government (MSIT) (No. RS-2019-II190079, Artificial Intelligence Graduate School Program, Korea University), and by the Information Technology Research Center (ITRC) support program (No. IITP-2026-RS-2024-00436857).}
\thanks{
$^{1}$G.-H. Na, H.-J. Kim, and S.-W. Lee are with the Department of Artificial Intelligence, Korea University, Anam-dong, Seongbuk-ku, Seoul 02841, Korea.
\texttt{\small \{g\_h\_na, hojoong\_kim, sw.lee\}@korea.ac.kr}
}}
\begin{document}

\maketitle
\thispagestyle{empty}
\pagestyle{empty}






\begin{abstract}
Zero-shot composed image retrieval (ZS-CIR) aims to retrieve a target image from a multimodal query consisting of a reference image and an edit text describing the desired modification. Recent ZS-CIR studies have relied on projection-based methods that map a reference image into pseudo-word tokens in the text embedding space. However, such methods require additional projection and re-encoding steps, increasing training complexity, reducing efficiency, and introducing a discrepancy between training and inference. In this paper, we propose DiCE-CIR, a direct composition learning method that predicts composed query representations by directly composing a reference image and an edit text. To enable scalable training without manually annotated triplets, we automatically construct compositional training samples from large-scale image-caption pairs using a large language model. Based on these samples, we train a lightweight composition module with objectives that promote alignment with the target, edit-consistent semantic transformation, and retrieval discriminability.
We conduct extensive experiments on ZS-CIR benchmarks and show that DiCE-CIR achieves state-of-the-art performance on CIRCO and competitive performance on CIRR while maintaining high computational efficiency.
\end{abstract}

\section{INTRODUCTION}
Composed image retrieval (CIR) aims to retrieve a target image from a multimodal query composed of a reference image and a textual instruction describing the modification~\cite{TIRG}. 
Compared with conventional image retrieval and visual matching, which typically rely on visual similarity or correspondence~\cite{Pill-ID, Nighttime, lee1995multilayer, roh2010view, lee1990translation}, CIR is challenging because jointly understanding the reference image and the modification specified by the text is required.

Early CIR research~\cite{Combiner,park2025far} has relied on supervised triplet datasets consisting of a reference image, an edit text, and a target image. Constructing such triplet annotations at scale is difficult and requires substantial manual effort.
Existing CIR datasets remain limited in scale (e.g., FashionIQ~\cite{FashionIQ} with 36.1k samples and CIRR~\cite{CIRR} with 36.5k samples), which restricts generalization across diverse domains.
 
To alleviate the limitations of supervised CIR, recent studies have turned to zero-shot CIR (ZS-CIR), which leverages the semantic alignment ability of pretrained vision-language models (VLMs) without requiring task-specific CIR supervision. Most ZS-CIR studies~\cite{pic2word,CIRCO_SEARLE,LinCIR,FTI4CIR} adopt projection-based methods built on CLIP~\cite{CLIP}, which map reference image features into the text embedding space and reformulate CIR as a text-to-image retrieval problem.
However, projection-based methods suffer from two structural limitations: (i) a sequential dependency during training caused by the additional projection and re-encoding steps, and (ii) a training-inference discrepancy, because training optimizes the mapping of the reference image into a pseudo-word token, rather than directly optimizing the composed query used for retrieval at inference.

\begin{figure}[t]
    \centering

    \begin{subfigure}[t]{0.94\columnwidth}
        \centering
        \includegraphics[width=\linewidth]{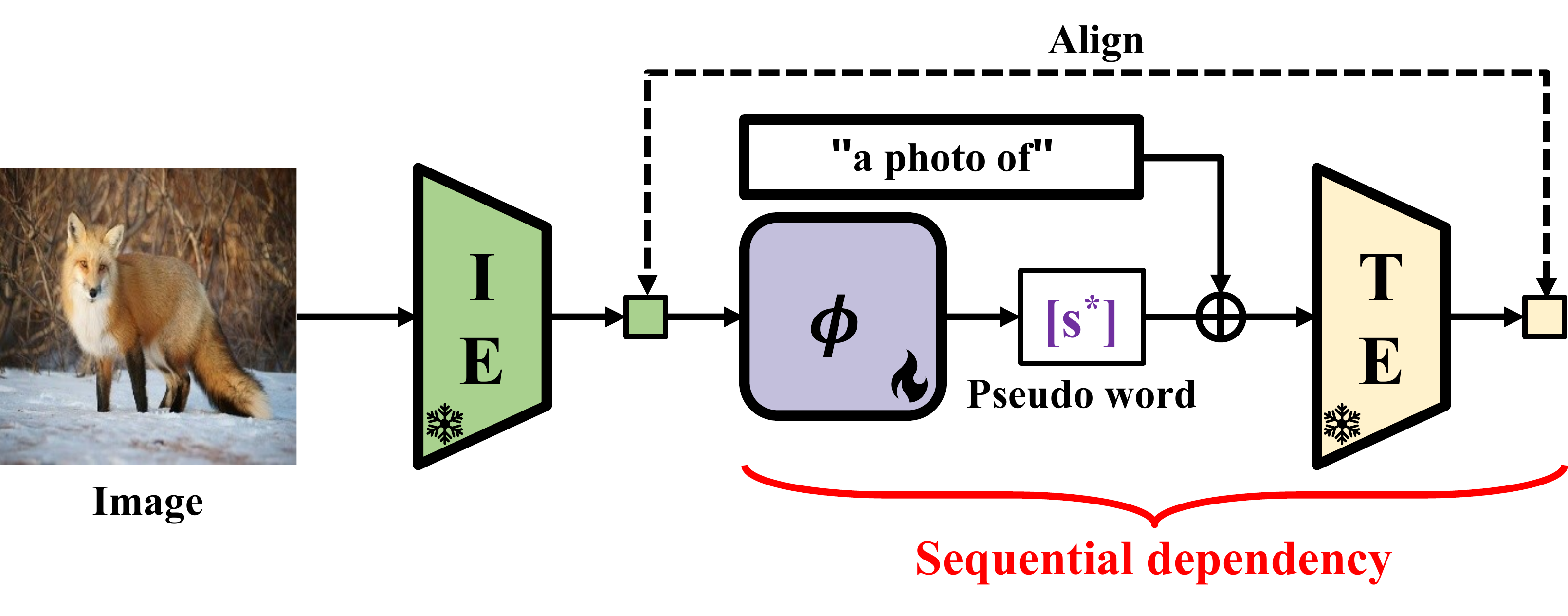}
        \vspace{-12pt}
        \caption{Sequential dependency during training.}
        \label{fig:fig1_a}
    \end{subfigure}


    \begin{subfigure}[t]{\columnwidth}
        \centering
        \includegraphics[width=\linewidth]{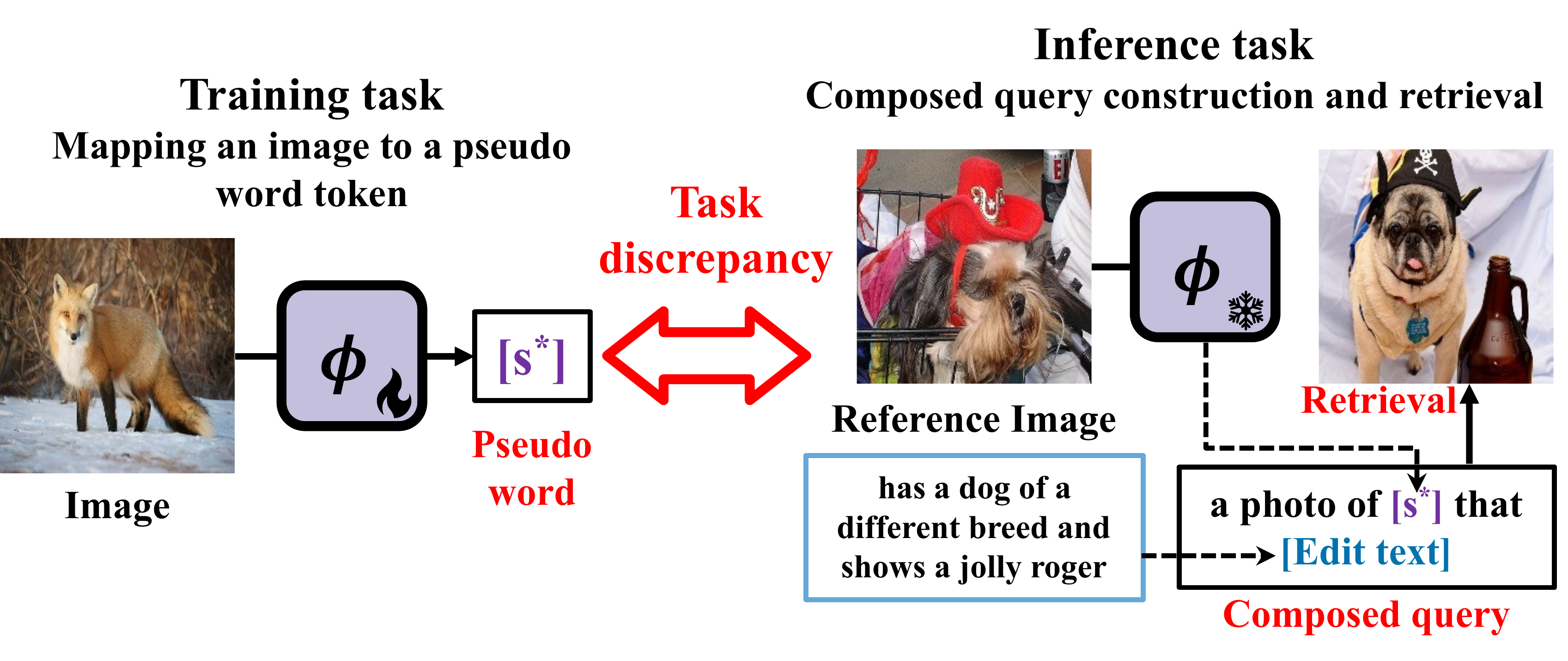}
        \vspace{-12pt}
        \caption{Training-inference discrepancy.}
        \label{fig:fig1_b}
    \end{subfigure}
    \caption{\textbf{Limitations of projection-based methods.} 
    (a) Sequential dependency during training in projection-based methods.
    (b) Training-inference discrepancy in projection-based methods.
    }
    \vspace{-10pt}
    
    \label{fig:fig1}
\end{figure}

Fig.~\ref{fig:fig1}(a) shows the sequential dependency of projection-based methods during training. These projection-based methods do not directly combine the reference image with the edit text. Instead, an additional projection module first projects the reference image into a pseudo-word token in the text embedding space. During training, the pseudo-word token is combined with a text prompt and then fed into the CLIP text encoder. Because pseudo-word token generation must precede CLIP text encoding, the pipeline becomes sequential, requiring additional projection and re-encoding steps that increase training complexity and reduce computational efficiency.
Fig.~\ref{fig:fig1}(b) shows the training-inference discrepancy of projection-based methods. During training, projection-based methods optimize the mapping of the reference image into the text embedding space as a pseudo-word token. In contrast, during inference, the pseudo-word token is combined with the edit text in a fixed prompt to form the composed query for retrieval. Training optimizes only the projection of the reference image into the text embedding space, rather than directly optimizing the composed query that reflects the intended modification. This discrepancy creates a gap between the training objective and the retrieval process at inference, making it difficult to form a composed query that is directly aligned with retrieval.

To address these limitations, we propose a direct composition learning for efficient zero-shot composed image retrieval (DiCE-CIR), which directly composes a reference image and an edit text to predict a composed query. 
To enable scalable training, we first construct compositional training samples automatically from large-scale image-caption pairs. Specifically, for each image in CC3M~\cite{CC3M}, we use the original caption as the reference caption and employ a large language model (LLM) to generate an edit text and a target caption. Each training sample consists of a reference image, a reference caption, an edit text, and a target caption, without requiring manual triplet annotation. 
This strategy enables large-scale image-caption corpora to serve as compositional supervision for CIR.
Because pretrained VLMs align image and text embeddings in a shared embedding space, reference and target captions can serve as scalable textual proxies for images during training.
Built on this data construction pipeline, DiCE-CIR is trained to combine a reference representation and an edit text using a lightweight composition module.
Unlike projection-based methods that convert the reference image into a pseudo-word token and feed it back into the text encoder, our method directly combines the two representations to produce a composed query.
This design simplifies the overall architecture and allows the same composition mechanism to be used during both training and inference.
Extensive evaluations on ZS-CIR benchmarks demonstrate that DiCE-CIR achieves state-of-the-art performance compared with existing projection-based methods, while also providing substantial gains in training efficiency. Under the same experimental setting, DiCE-CIR trains approximately 10.5$\times$ faster than LinCIR~\cite{LinCIR} with CLIP ViT-L and approximately 24.3$\times$ faster with CLIP ViT-G. These results show that DiCE-CIR achieves both strong retrieval performance and high training efficiency.

The main contributions of this paper are as follows:
\begin{itemize}
    \item We propose DiCE-CIR, a direct composition method for ZS-CIR, with an automatic compositional data construction strategy that converts large-scale image-caption pairs into compositional training samples.
    \item We demonstrate that direct composition of a reference image and an edit-text alleviates the sequential dependency during training and the training-inference discrepancy of projection-based methods.
    \item Extensive experiments on ZS-CIR benchmarks demonstrate that DiCE-CIR achieves state-of-the-art performance with substantially improved training efficiency.
\end{itemize}     
\section{RELATED WORK}
\subsection{Zero-Shot Composed Image Retrieval}
ZS-CIR aims to remove the dependence on costly triplet annotations by exploiting pretrained VLMs.
Most existing studies adopt projection-based methods that map a reference image into a pseudo-word token in the CLIP embedding space and combine the token with the edit text to construct a composed query.
Pic2Word~\cite{pic2word} first introduced a projection-based method by reformulating CIR as a text-to-image retrieval problem.
SEARLE~\cite{CIRCO_SEARLE} enhances pseudo-word expressiveness by obtaining pseudo-word tokens through optimization-based textual inversion and distilling the pseudo-word tokens into a textual inversion network.
FTI4CIR~\cite{FTI4CIR} extends the single-token setting by introducing subject-oriented and attribute-oriented pseudo-word tokens.
LinCIR~\cite{LinCIR} improves efficiency and scalability by training the projection module with language-only self-supervision, without image supervision. 
CIG~\cite{CIG} generates a pseudo target image to provide additional visual cues for composed query construction.
PrediCIR~\cite{predcir} leverages world-model-based prediction to complement target-relevant visual content missing from the reference image in the latent space and incorporates the predicted content into pseudo-word mapping.
In contrast, our method learns a composed query by directly composing a reference representation and an edit-text without pseudo-word projection.

\subsection{Vision-Language Models}
VLMs have achieved strong performance on various cross-modal tasks by aligning images and texts in a shared embedding space through large-scale image-text pretraining. 
CLIP~\cite{CLIP} is a representative model that enables semantic alignment between visual and textual embeddings. This shared embedding space allows direct comparison across modalities, which is fundamental to vision-language retrieval.
Moreover, the shared embedding space enables the use of textual embeddings that reflect target semantics as supervision signals, even without explicit access to the target image.
Building on this property, our method directly composes a reference embedding and an edit-text embedding in the shared embedding space induced by the VLM to learn a composed query embedding.
\begin{figure*}[t]
    \centering
    \includegraphics[width=0.95\linewidth]{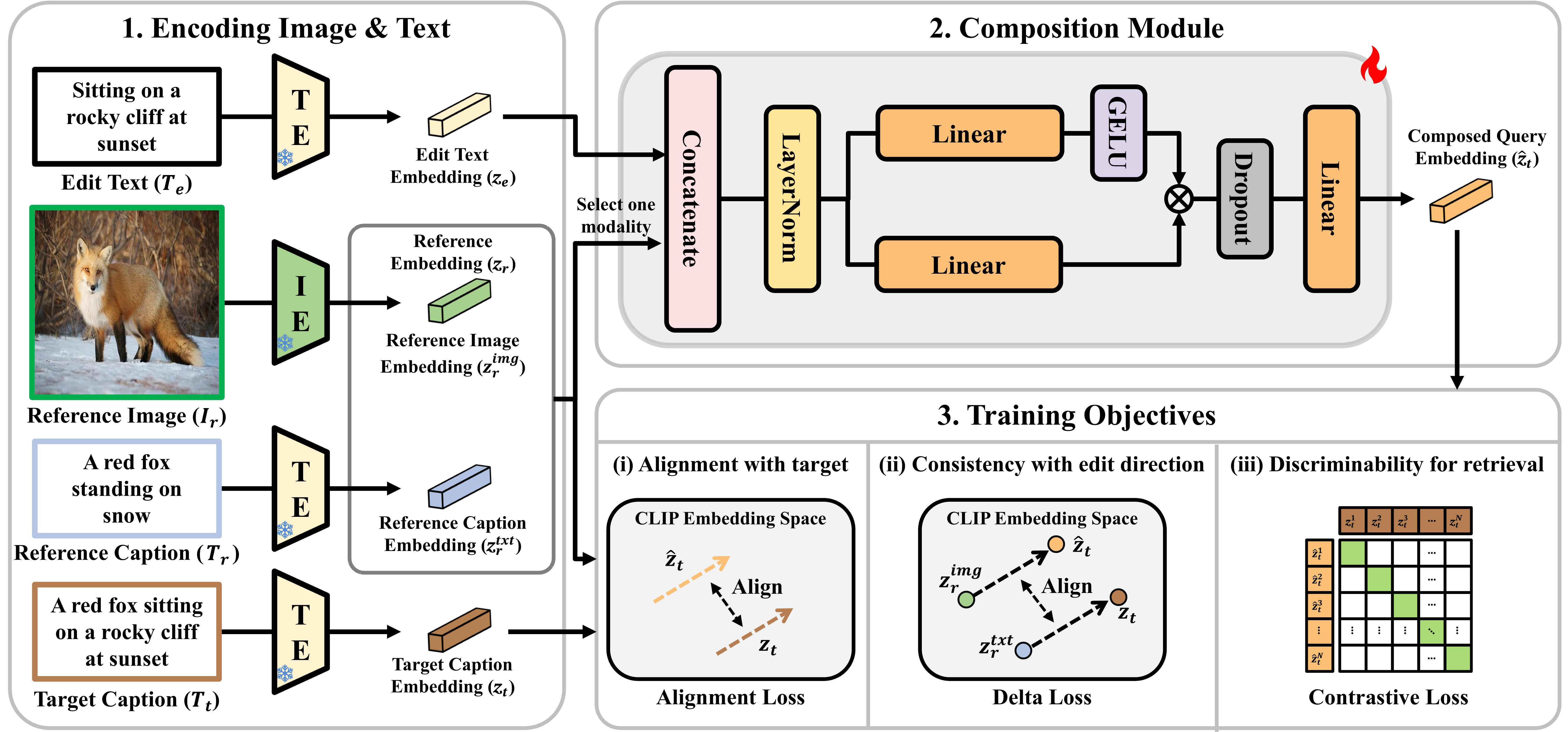}
    \vspace{-4pt}
    \caption{\textbf{Overview of DiCE-CIR.}
    Given a training sample $(I_r, T_r, T_e, T_t)$, DiCE-CIR encodes the reference image, reference caption, edit text, and target caption into a shared embedding space using frozen CLIP encoders.
    A lightweight composition module combines a reference embedding $z_r \in \{z_r^{\mathrm{img}}, z_r^{\mathrm{txt}}\}$ with the edit-text embedding $z_e$ to predict the composed query embedding $\hat{z}_t = \Phi(z_r, z_e)$.
    Our method is trained to align $\hat{z}_t$ with the target caption embedding $z_t$ while enforcing edit-consistent semantic transformation and retrieval discriminability.}
    \label{fig:method_overview}
    \vspace{-15pt}
\end{figure*}
\section{METHOD}
\subsection{Preliminaries}
ZS-CIR aims to retrieve a target image from an image database given a multimodal query consisting of a reference image and an edit text, such that the retrieved image reflects the requested modification while preserving the relevant semantics of the reference image. Formally, let $I_r$ denote the reference image, $T_e$ denote the edit text describing the desired modification, and $\mathcal{D}=\{I_i\}_{i=1}^{N}$ denote the image database of $N$ candidate images. Given the query $(I_r, T_e)$, the objective is to rank the target image $I_t \in \mathcal{D}$ within the top-$K$ retrieved results.
Following prior ZS-CIR methods, we adopt a pretrained CLIP~\cite{CLIP,openclip} model as the backbone encoder. We denote the frozen CLIP image encoder and text encoder by $f_{\mathrm{img}}$ and $f_{\mathrm{txt}}$, respectively, which map visual and textual inputs into a shared embedding space. On top of these frozen encoders, the only trainable component in our method is a lightweight composition module that directly combines the reference image embedding and edit-text embedding.

\subsection{Automatic Construction of Compositional Training Data}
Conventional CIR training has largely relied on manually annotated triplets consisting of a reference image, an edit text, and a target image. However, such supervision is costly to obtain and difficult to scale. To alleviate this limitation, we automatically construct compositional training samples for ZS-CIR from the large-scale image-caption dataset CC3M~\cite{CC3M}.
For each image-caption pair in CC3M, we use the original image as the reference image and its associated caption as the reference caption. We then provide an LLM with the reference caption, a predefined edit type, and a prompt template to generate an edit text and a target caption.
The edit text describes only the visual changes with respect to the reference, while the target caption provides a standalone description of the edited target image.
To increase the diversity of generated edits, we define five edit types: \textit{fine-grained attribute}, \textit{spatial or viewpoint}, \textit{strong scene change}, \textit{compositional two change}, and \textit{retrieval short}. The prompt template used for LLM-based compositional data construction is shown in Fig.~\ref{fig:data_prompt}.
Each training sample is represented as $(I_r, T_r, T_e, T_t)$, where $I_r$ is the reference image, $T_r$ is the reference caption, $T_e$ is the edit text, and $T_t$ is the target caption. We construct approximately 1.22M compositional training data from CC3M.
This strategy provides scalable compositional supervision for learning compositional semantic transformations without requiring manual CIR triplet annotations. This motivation is also consistent with prior studies in limited-supervision visual recognition, where improving generalization from scarce annotations has been a central objective~\cite{jung2020few, seo2021self, kim2021spatial, kim2018discriminative}.


\begin{figure}[t]
\centering
\begin{tcolorbox}[
    enhanced,
    width=0.94\linewidth,
    colback=gray!8,
    colframe=black!70,
    boxrule=0.7pt,
    arc=3mm,
    outer arc=3mm,
    left=4mm,
    right=4mm,
    top=2.5mm,
    bottom=2.5mm,
    title={\textbf{Data Generation Prompt}},
    colbacktitle=black!75,
    coltitle=white,
    fonttitle=\bfseries,
    attach boxed title to top left={xshift=0mm,yshift=0mm},
    boxed title style={
        size=small,
        boxrule=0pt,
        colframe=black!75,
        colback=black!75,
        sharp corners
    }
    ]
\small
\rmfamily
\raggedright

\textbf{Input.}
A reference caption $\{T_r\}$ and a target edit type $\{E\}$.

\vspace{0.4em}
\textbf{Output.}
Generate exactly one edited target example consisting of
\textit{edit type}, \textit{edit text}, and \textit{target caption}.

\vspace{0.4em}
\textbf{Constraints.}
\begin{itemize}
    \item Preserve the same main subject.
    \item Let \textit{edit text} describe only the visual difference from the reference.
    \item Let \textit{target caption} be a full standalone caption.
    \item Keep the result natural and visually plausible.
    \item Ensure that \textit{edit type} exactly matches $\{E\}$.
    \item If $\{E\}$ is \textit{compositional two change}, include exactly two visible changes.
    \item If $\{E\}$ is \textit{retrieval short}, keep the edit concise.
\end{itemize}

\end{tcolorbox}
\vspace{-8pt}
\caption{\textbf{Prompt template used for LLM-based compositional data construction.} Our method uses only text captions for automatic compositional data construction.}
\label{fig:data_prompt}
\vspace{-12pt}
\end{figure}
\subsection{Compositional Embedding Learning}
An overview of the proposed method is illustrated in Fig.~\ref{fig:method_overview}. 
Given a training sample $(I_r, T_r, T_e, T_t)$, we encode the reference image, reference caption, edit text, and target caption using frozen CLIP encoders:
\begin{equation}
\begin{split}
\mathbf{z}_{r}^{\mathrm{img}} = f_{\mathrm{img}}(I_r), \quad
\mathbf{z}_{r}^{\mathrm{txt}} = f_{\mathrm{txt}}(T_r), \\
\mathbf{z}_{e} = f_{\mathrm{txt}}(T_e), \quad
\mathbf{z}_{t} = f_{\mathrm{txt}}(T_t),
\end{split}
\end{equation}
where all embeddings lie in a shared $D$-dimensional space.

\subsubsection{Composition Module}
We predict a composed query embedding by composing a reference embedding with the edit-text embedding:
\begin{equation}
\hat{\mathbf{z}}_{t} =
\Phi(\mathbf{z}_{r}, \mathbf{z}_{e}), \quad
\mathbf{z}_{r} \in \left\{\mathbf{z}_{r}^{\mathrm{img}}, \mathbf{z}_{r}^{\mathrm{txt}}\right\}\text{,}
\end{equation}
where $\mathbf{z}_{r}$ denotes a reference embedding, which can be either the reference image embedding $\mathbf{z}_{r}^{\mathrm{img}}$ or the reference caption embedding $\mathbf{z}_{r}^{\mathrm{txt}}$. During training, each sample is expanded into two compositional training instances by pairing the edit-text embedding with both reference modalities.
We instantiate $\Phi$ as a gated MLP operating on the concatenated reference embedding and edit-text embedding. We first concatenate the reference embedding and the edit-text embedding, and then apply layer normalization:
\begin{equation}
\mathbf{x} = \mathrm{LayerNorm}\big([\mathbf{z}_{r} ; \mathbf{z}_{e}]\big).
\end{equation}
The normalized embedding is passed through two parallel linear projections, one with GELU activation, followed by a linear projection to obtain the composed query embedding:
\begin{equation}
\hat{\mathbf{z}}_{t} = W_{p}\Big(\mathrm{GELU}(W_{1}\mathbf{x}) \odot (W_{2}\mathbf{x})\Big).
\end{equation}

\subsubsection{Training Objectives}
We train $\Phi$ with objectives that encourage three properties: (i) alignment with the target caption embedding, (ii) consistency with the intended edit direction, and (iii) discriminability for retrieval.

First, we use an alignment loss to pull the composed query embedding toward the target caption embedding:
\begin{equation}
\mathcal{L}_{\mathrm{align}} = 1 - \mathrm{sim}(\hat{\mathbf{z}}_{t}, \mathbf{z}_{t}),
\end{equation}
where $\mathrm{sim}(\cdot,\cdot)$ denotes cosine similarity. This alignment loss provides a direct supervision signal that encourages the composed query embedding to capture the global semantics of the target caption.

Aligning the composed query embedding with the target caption embedding alone does not explicitly model how the target should differ from the reference. To capture the semantic transformation induced by the edit text, we align the residual from the reference image embedding to the composed query embedding with the residual from the reference caption embedding to the target caption embedding:
\begin{equation}
\mathcal{L}_{\mathrm{delta}} =
1 - \mathrm{sim}\big(
\hat{\mathbf{z}}_{t} - \mathbf{z}_{r}^{\mathrm{img}},
\mathbf{z}_{t} - \mathbf{z}_{r}^{\mathrm{txt}}
\big).
\end{equation}
By aligning these residual directions, our method is encouraged to approach the target semantics while following an edit-consistent semantic shift relative to the reference.

To make the learned embedding suitable for retrieval, we employ in-batch contrastive learning. The contrastive loss encourages each composed query embedding to be close to its matched target caption embedding while being separated from other target caption embeddings in the batch:
\begin{equation}
\mathcal{L}_{\mathrm{ctr}} =
-\log
\frac{
\exp\!\left(\mathrm{sim}(\hat{\mathbf{z}}_{t}, \mathbf{z}_{t})/\tau\right)
}{
\sum_{j} \exp\!\left(\mathrm{sim}(\hat{\mathbf{z}}_{t}, \mathbf{z}_{j})/\tau\right)
}\text{,}
\end{equation}
where $\mathbf{z}_{j}$ denotes the target caption embedding of the $j$-th sample in the batch. This contrastive loss improves discriminability by encouraging the composed query embedding to identify the correct target caption embedding against competing negatives.

The final objective is defined as:
\begin{equation}
\mathcal{L} =
\lambda_{\mathrm{align}}\mathcal{L}_{\mathrm{align}}
+ \lambda_{\mathrm{delta}}\mathcal{L}_{\mathrm{delta}}
+ \lambda_{\mathrm{ctr}}\mathcal{L}_{\mathrm{ctr}}.
\end{equation}
Overall, the objective encourages our method to learn a composed query embedding that is target-aware, edit-consistent, and discriminative for retrieval.
\subsection{Inference}
At inference, given a reference image $I_r$ and an edit text $T_e$, we encode them using the same frozen CLIP encoders as in training and compose their embeddings with $\Phi$ to form the composed query embedding for retrieval. 
We then compute the cosine similarity between the composed query embedding and each candidate image embedding in the database, and return the top-$K$ ranked results.
Because the same composition module $\Phi$ is used during both training and inference, our method mitigates the training-inference discrepancy of prior projection-based methods.           
\begin{table*}[t]
\centering
\caption{\textbf{Performance comparison on CIRCO and CIRR.} The best and second-best results are highlighted in bold and underline, respectively. Our method is highlighted in gray.}
\vspace{-5pt}
\label{table:main}
\resizebox{0.88\linewidth}{!}{
\begin{tabular}{llcccc c cccccc}
\toprule
\multirow{2.5}{*}{\textbf{Backbone}} & \multirow{2.5}{*}{\textbf{Method}}
& \multicolumn{4}{c}{\textbf{CIRCO}}
&
& \multicolumn{6}{c}{\textbf{CIRR}} \\
\cmidrule(r{0.1em}){3-6}
\cmidrule(l{0.1em}){8-13}
& & \textbf{mAP@5} & \textbf{mAP@10} & \textbf{mAP@25} & \textbf{mAP@50}
& & \textbf{R@1} & \textbf{R@5} & \textbf{R@10} & \textbf{R$_s$@1} & \textbf{R$_s$@2} & \textbf{R$_s$@3} \\
\midrule

\multirow{10}{*}{\textbf{ViT-L/14}}
& Image-only                 & 2.53 & 2.89 & 3.60 & 3.96 & & 7.21 & 23.28 & 33.88 & 20.41 & 41.45 & 61.08 \\
& Text-only                  & 3.10 & 3.29 & 3.80 & 4.03 & & 21.61 & 46.77 & 58.79 & \textbf{62.24} & \textbf{80.60} & \underline{90.53} \\
& Image+Text                 & 7.21 & 7.96 & 9.04 & 9.59 & & 13.49 & 37.66 & 51.13 & 34.79 & 59.32 & 76.46 \\
& Pic2Word~\cite{pic2word}   & 8.72 & 9.51 & 10.64 & 11.29 & & 23.90 & 51.70 & 65.30 & 53.76 & 74.46 & 87.08 \\
& SEARLE~\cite{CIRCO_SEARLE} & 11.68 & 12.73 & 14.33 & 15.12 & & 24.24 & 52.48 & 66.29 & 53.76 & 75.01 & 88.19 \\
& LinCIR~\cite{LinCIR}       & 12.59 & 13.58 & 15.00 & 15.85 & & 25.04 & 53.25 & 66.68 & 57.11 & 77.37 & 88.89 \\
& FTI4CIR~\cite{FTI4CIR}     & 15.05 & 16.32 & 18.06 & 19.05 & & 25.90 & 55.61 & 67.66 & 55.21 & 75.88 & 87.98 \\
& CIG~\cite{CIG}             & 12.97 & 13.64 & 15.14 & 16.01 & & 26.72 & 55.52 & 68.10 & 57.95 & 77.81 & 89.45 \\
& PrediCIR~\cite{predcir}    & \underline{15.70} & \underline{17.10} & \underline{18.60} & \underline{19.30} & & \underline{27.20} & \underline{57.00} & \underline{70.20} & - & - & - \\
& \cellcolor{gray!15}\textbf{DiCE-CIR}
  & \cellcolor{gray!15}\textbf{20.78}
  & \cellcolor{gray!15}\textbf{21.66}
  & \cellcolor{gray!15}\textbf{23.75}
  & \cellcolor{gray!15}\textbf{24.67}
  & \cellcolor{gray!15}
  & \cellcolor{gray!15}\textbf{28.22}
  & \cellcolor{gray!15}\textbf{57.95}
  & \cellcolor{gray!15}\textbf{70.46}
  & \cellcolor{gray!15}\underline{61.37}
  & \cellcolor{gray!15}\underline{79.69}
  & \cellcolor{gray!15}\textbf{90.77} \\
\midrule

\multirow{7}{*}{\textbf{ViT-G/14}}
& Image-only                 & 3.87 & 4.21 & 5.02 & 5.53 & & 8.00 & 25.69 & 36.53 & 20.75 & 40.82 & 60.60 \\
& Text-only                  & 4.70 & 5.04 & 5.69 & 6.00 & & 29.69 & 56.05 & 67.23 & \textbf{69.57} & \textbf{85.88} & \textbf{93.40} \\
& LinCIR~\cite{LinCIR}       & 19.71 & 21.01 & 23.13 & 24.18 & & 35.25 & 64.72 & 76.05 & 63.35 & 82.22 & 91.98 \\
& CIG~\cite{CIG}             & 20.64 & 21.90 & 24.04 & 25.20 & & \underline{36.05} & \underline{66.31} & 76.96 & 64.94 & 83.18 & 91.93 \\
& PrediCIR~\cite{predcir}    & \underline{23.70} & \underline{24.60} & \underline{25.40} & \underline{26.00} & & \textbf{37.00} & 66.10 & \underline{77.90} & - & - & - \\
& \cellcolor{gray!15}\textbf{DiCE-CIR}
  & \cellcolor{gray!15}\textbf{26.23}
  & \cellcolor{gray!15}\textbf{27.52}
  & \cellcolor{gray!15}\textbf{29.82}
  & \cellcolor{gray!15}\textbf{30.88}
  & \cellcolor{gray!15}
  & \cellcolor{gray!15}35.88
  & \cellcolor{gray!15}\textbf{67.04}
  & \cellcolor{gray!15}\textbf{78.19}
  & \cellcolor{gray!15}\underline{66.48}
  & \cellcolor{gray!15}\underline{84.05}
  & \cellcolor{gray!15}\underline{92.46} \\
\bottomrule
\end{tabular}
}
\vspace{-14pt}
\end{table*}
\section{EXPERIMENTS}
\subsection{Experimental Setup}
\subsubsection{Baselines}
We compare the proposed method with prior projection-based methods. For fair comparison, we use two pretrained CLIP backbones, OpenAI CLIP ViT-L/14~\cite{CLIP} and OpenCLIP ViT-G/14~\cite{openclip}, and report results for each backbone separately. Pic2Word~\cite{pic2word}, SEARLE~\cite{CIRCO_SEARLE}, and LinCIR~\cite{LinCIR} are based on three-layer MLPs and are trained on 3M images, 100K images, and 5.5M text samples, respectively. FTI4CIR~\cite{FTI4CIR} uses an MLP with a small Transformer and is trained on 100K images. CIG~\cite{CIG} is built on a latent diffusion model and trained on 490K image-caption pairs, whereas PrediCIR~\cite{predcir} uses a 12-layer Transformer-based predictor and is trained on 3M image-caption pairs.

\subsubsection{Datasets and Evaluation Metrics}
We evaluate the proposed method on two standard ZS-CIR benchmarks, CIRR~\cite{CIRR} and CIRCO~\cite{CIRCO_SEARLE}, following their official evaluation protocols. CIRR evaluates the standard CIR setting, where the goal is to retrieve a ground-truth target image given a reference image and an edit text. In contrast, CIRCO reflects a more open-ended retrieval scenario in which multiple images can be relevant to a given query. Following prior work~\cite{LinCIR,FTI4CIR}, we report recall at K (R@K) and subset recall at K (R$_s$@K) for CIRR, and mAP@K for CIRCO.

\subsubsection{Implementation Details}
For both backbones, the image and text encoders are frozen, and only the composition module $\Phi$ is trained. The embedding dimension $D$ is 768 for ViT-L/14 and 1280 for ViT-G/14. We implement $\Phi$ as a gated MLP with hidden dimension $4D$ and dropout rate 0.05. We use 1.22M compositional training data automatically constructed from CC3M~\cite{CC3M} using Qwen3.5-9B~\cite{Qwen3.5}. We optimize DiCE-CIR with AdamW using a learning rate of 0.005, weight decay of 0.05, and a constant learning rate schedule with 4,000 warmup steps over 20,000 training steps. The batch size is 512. The loss weights are $\lambda_{\mathrm{align}}=1.0$, $\lambda_{\mathrm{delta}}=1.0$, and $\lambda_{\mathrm{ctr}}=0.2$, and the contrastive temperature is $\tau=0.07$. We apply exponential moving average (EMA) with a decay factor of 0.9995 during training. All experiments are conducted on two NVIDIA RTX A6000 GPUs.

\subsection{Main Results}
\subsubsection{CIRCO}
Tab.~\ref{table:main} shows the performance comparison on the CIRCO test set. DiCE-CIR consistently outperforms prior methods across all mAP metrics with both ViT-L/14 and ViT-G/14 backbones. CIRCO reflects a more open-ended retrieval setting with multiple valid targets per query, requiring the composed query representation to capture the intended composition while supporting accurate ranking. The consistent gains across all mAP metrics suggest that the proposed method learns composed query representations that generalize well to this open-ended retrieval setting.

\subsubsection{CIRR}
Tab.~\ref{table:main} also shows the performance comparison on the CIRR test set. DiCE-CIR outperforms prior methods on all R@K metrics with ViT-L/14 and on R@5 and R@10 with ViT-G/14. In contrast, the text-only baseline remains relatively strong on R$_s$@K, which evaluates retrieval over a smaller image subset. 
As noted in prior work~\cite{CIRCO_SEARLE,LinCIR}, the edit texts in CIRR are not always described as consistent relative changes with respect to the reference image. In some cases, the edit text alone provides sufficient semantic cues to identify the target image, and incorporating the reference image may even interfere with retrieval. 
This tendency can be more pronounced for R$_s$@K, which is evaluated on a smaller candidate subset. 
Nevertheless, DiCE-CIR remains competitive on R$_s$@K compared with prior projection-based methods, suggesting that it learns effective composed query representations on CIRR as well.

\begin{table}[t]
\centering
\caption{\textbf{Ablation study on CIRR and CIRCO.}}
\vspace{-5pt}
\label{tab:ablation}
\setlength{\tabcolsep}{6pt}
\renewcommand{\arraystretch}{1.12}
\resizebox{0.95\linewidth}{!}{
\begin{tabular}{lccccc}
\toprule
\multirow{2}{*}{\textbf{Method}} & \multicolumn{3}{c}{\textbf{CIRR}} & \multicolumn{2}{c}{\textbf{CIRCO}} \\
\cmidrule(lr){2-4} \cmidrule(lr){5-6}
& \textbf{R@1} & \textbf{R@5} & \textbf{R@10} & \textbf{mAP@5} & \textbf{mAP@10} \\
\midrule

\textbf{Full method} & 28.22 & 57.95 & 70.46 & 20.78 & 21.66 \\
\midrule
\addlinespace[1pt]
\textbf{A. Training objectives} & & & & & \\
\quad A.1 w/o $\mathcal{L}_{\mathrm{align}}$ & 27.47 & 57.01 & 70.55 & 20.50 & 21.56 \\
\quad A.2 w/o $\mathcal{L}_{\mathrm{delta}}$ & 26.29 & 56.65 & 69.81 & 19.68 & 20.90 \\
\quad A.3 w/o $\mathcal{L}_{\mathrm{ctr}}$ & 25.78 & 55.11 & 66.51 & 15.79 & 16.48 \\
\midrule
\addlinespace[1pt]
\textbf{B. Architecture and training} & & & & & \\
\quad B.1 Three-layer MLP & 26.15 & 56.60 & 69.30 & 18.71 & 19.57 \\
\quad B.2 w/o EMA & 27.01 & 56.99 & 69.61 & 19.06 & 19.81 \\
\midrule
\addlinespace[1pt]
\textbf{C. Reference input} & & & & & \\
\quad C.1 Reference image only & 27.25 & 57.49 & 69.74 & 19.68 & 20.42 \\
\quad C.2 Reference caption only & 26.67 & 55.52 & 67.95 & 19.82 & 20.85 \\

\bottomrule
\end{tabular}
}
\vspace{-16pt}
\end{table}
\subsection{Ablation Studies}
\subsubsection{Effect of Training Objectives}
Tab.~\ref{tab:ablation} A shows the effect of each objective term. Removing any objective term degrades performance on both CIRR and CIRCO, confirming that all three objectives contribute to the final retrieval performance. In particular, removing $\mathcal{L}_{\mathrm{delta}}$ leads to a clear drop on both benchmarks, indicating that modeling the semantic transformation from the reference to the target is a key component of composed image retrieval. The largest degradation is observed when $\mathcal{L}_{\mathrm{ctr}}$ is removed, showing that the contrastive objective is critical for learning retrieval-discriminative composed query embeddings. Removing $\mathcal{L}_{\mathrm{align}}$ also reduces performance, suggesting that direct alignment with the target caption embedding provides complementary supervision.

\subsubsection{Effect of Architecture and Training}
Tab.~\ref{tab:ablation} B shows the effect of the architecture and training setup. Replacing the gated MLP with a three-layer MLP degrades performance on both CIRR and CIRCO, indicating that the gated design is more effective for composition. Notably, the gated MLP also uses substantially fewer parameters than the three-layer MLP (11.81M vs. 16.53M for ViT-L/14, and 32.78M vs. 45.89M for ViT-G/14), suggesting a better parameter-efficiency. EMA provides additional gains, by improving training stability.

\subsubsection{Effect of Reference Input}
Tab.~\ref{tab:ablation} C shows the effect of the reference input. Using both reference modalities yields the best performance on both CIRR and CIRCO, indicating that the reference image and reference caption provide complementary information for composition. Even when using a single reference modality, our method still achieves competitive performance.

\subsection{Further Analysis}
\subsubsection{Efficiency Analysis}
Tab.~\ref{tab:efficiency} summarizes the efficiency comparison between DiCE-CIR and prior ZS-CIR methods. DiCE-CIR enables faster training and inference across backbones. This efficiency gain comes from the proposed direct composition method, which avoids the projection and re-encoding steps required by projection-based methods. By directly composing a reference embedding and an edit-text embedding without pseudo-word re-encoding, DiCE-CIR removes the sequential dependency in projection-based pipelines and enables efficient training and inference.

\subsubsection{Qualitative Analysis}
\begin{figure}[t]
    \centering
    \vspace{2.6mm}
    \includegraphics[width=\linewidth]{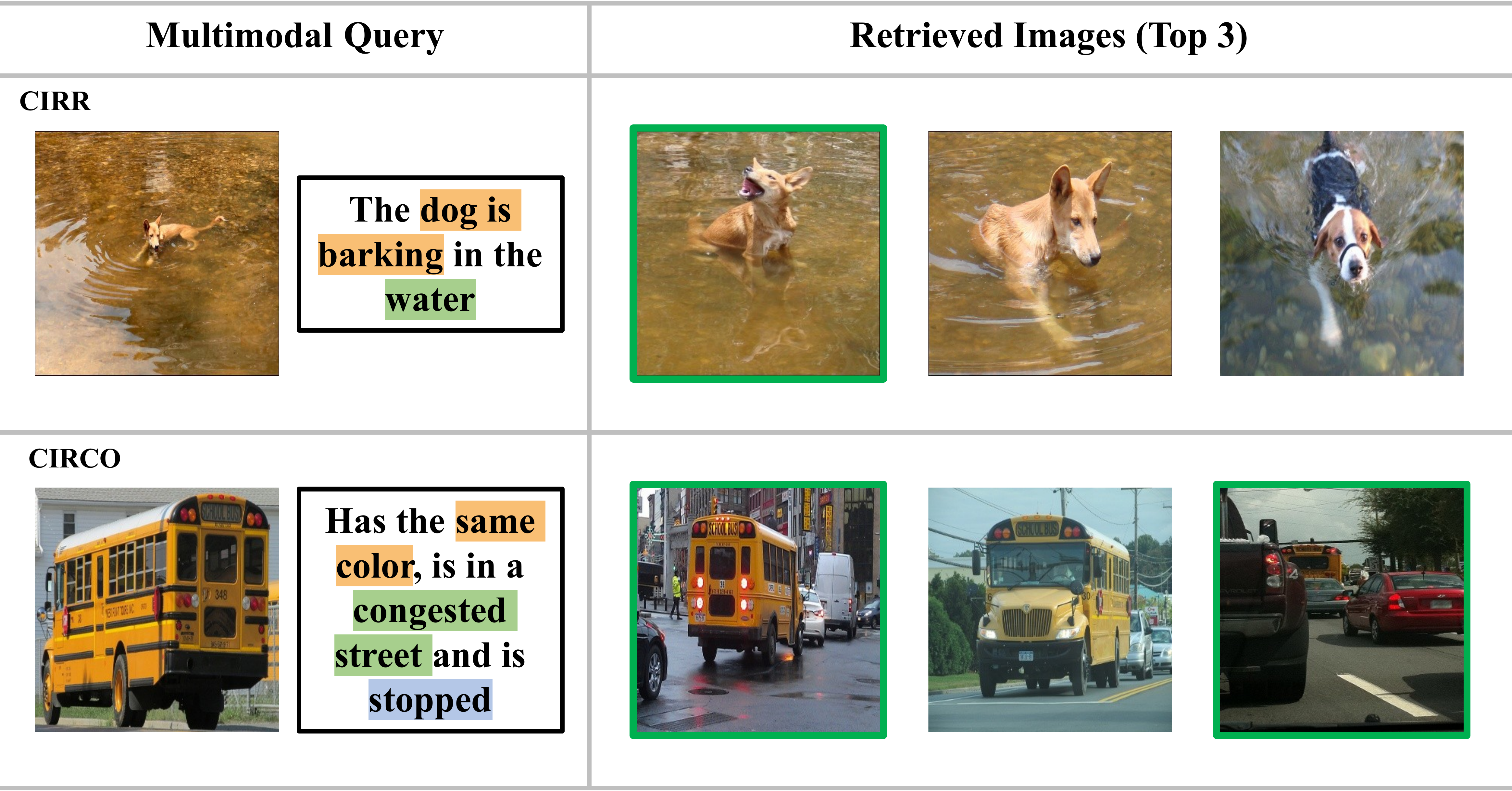}
    \vspace{-18pt}
    \caption{\textbf{Qualitative retrieval results on CIRR and CIRCO.} The figure shows representative success cases of DiCE-CIR on the two ZS-CIR benchmarks.}
    \label{fig:qualitative_results}
    \vspace{-16pt}
\end{figure}

Fig.~\ref{fig:qualitative_results} shows that DiCE-CIR retrieves images consistent with the edit text while preserving the key visual content of the reference image. These examples suggest that DiCE-CIR performs well not only for relatively simple edit requests in CIRR but also for more compositional edit requests with multiple semantic constraints in CIRCO.

\begin{table}[t]
\centering
\caption{\textbf{Training and inference time comparison.} Training and inference times are reported in hours and seconds, respectively. Methods marked with $^\dagger$ are evaluated under the same environment as DiCE-CIR; the others follow the numbers reported in the original papers.}
\label{tab:efficiency}
\vspace{-5pt}
\setlength{\tabcolsep}{5pt}
\renewcommand{\arraystretch}{1.12}
\resizebox{0.95\linewidth}{!}{
\begin{tabular}{clccc}
\toprule
\textbf{Backbone} & \textbf{Method} & \textbf{Training (h)} & \textbf{Inference (s)} & \textbf{Training GPUs} \\
\midrule

\multirow{6}{*}{\textbf{ViT-L}}
& Pic2Word         & 3.0  & 0.02  & A100 $\times$ 8 \\
& SEARLE           & 4.2  & 0.02  & A100 $\times$ 8 \\
& LinCIR           & 0.5  & 0.02  & A100 $\times$ 8 \\
& LinCIR$^\dagger$ & 2.1  & 0.05    & RTX A6000 $\times$ 2 \\
& PrediCIR         & 28.0 & 0.12  & A100 $\times$ 4 \\
& \cellcolor{gray!15}\textbf{DiCE-CIR} & \cellcolor{gray!15}\textbf{0.2} & \cellcolor{gray!15}0.01 & \cellcolor{gray!15}RTX A6000 $\times$ 2 \\
\midrule

\multirow{5}{*}{\textbf{ViT-G}}
& Pic2Word         & 13.4 & 0.050 & A100 $\times$ 8 \\
& SEARLE           & 14.4 & 0.047 & A100 $\times$ 8 \\
& LinCIR           & 0.8  & 0.047 & A100 $\times$ 8 \\
& LinCIR$^\dagger$ & 7.3   & 0.095    & RTX A6000 $\times$ 2 \\
& \cellcolor{gray!15}\textbf{DiCE-CIR} & \cellcolor{gray!15}\textbf{0.3} & \cellcolor{gray!15}0.020 & \cellcolor{gray!15}RTX A6000 $\times$ 2 \\
\bottomrule
\end{tabular}
}
\vspace{-15pt}
\end{table}

\section{CONCLUSION}

In this paper, we presented DiCE-CIR, a direct composition method for efficient zero-shot composed image retrieval. DiCE-CIR automatically constructs compositional training data from large-scale image-caption pairs using an LLM and learns a lightweight module that directly combines a reference image and an edit text into a composed query. Experiments demonstrate that DiCE-CIR achieves strong retrieval performance with improved efficiency, validating the effectiveness of direct composition learning.





\addtolength{\textheight}{-12cm}   

\bibliographystyle{IEEEtran}
\bibliography{REFERENCE} 
\end{document}